\documentclass[letterpaper, 10 pt, conference]{ieeeconf}  

\IEEEoverridecommandlockouts
\usepackage{xurl}
\usepackage{balance}
\usepackage{amsmath}
\usepackage{amssymb,bm}
\usepackage{amsfonts}
\usepackage{fancyvrb}
\usepackage{enumerate}
\usepackage{tabulary}
\usepackage{multirow}
\usepackage{cite}
\usepackage{lipsum}  
\usepackage{graphicx}
\usepackage{epstopdf}
\usepackage[misc]{ifsym}
\usepackage{color}
\usepackage{xcolor}
\usepackage{xxcolor}
\usepackage{pgf}
\usepackage{pgfplots}
\usepgfplotslibrary{statistics}
\usepackage{tikz}
\usepackage{tikz-3dplot}
\usepackage{float}
\usepackage{lipsum}
\usepackage{booktabs}
\usepackage{verbatim}
\usepackage{hyperref}
\usepackage{makecell}
\usepackage{xfrac}
\usepackage{siunitx}
\usepackage{colortbl}
\usepackage{forest}

\usetikzlibrary{external, shapes.geometric, arrows, positioning, quotes, automata, backgrounds}
\tikzstyle{block} = [rectangle, rounded corners, minimum width=2.5cm, minimum height=1cm,text centered, draw=black, fill=blue!20]
\tikzstyle{red_block} = [rectangle, rounded corners, minimum width=2.5cm, minimum height=1cm,text centered, draw=black, fill=red!20]
\tikzstyle{inner_block} = [rectangle, rounded corners, minimum width=1.5cm, text centered, draw=black, fill=blue!20]
\tikzstyle{vision} = [rectangle, rounded corners, minimum width=2.5cm, text centered, draw=black, fill=green!20]
\tikzstyle{processor} = [rectangle, rounded corners, minimum width=2.5cm, minimum height=1cm, text centered, draw=black, fill=orange!20]
\tikzstyle{arrow} = [thick,->,>=stealth]
\tikzstyle{doublearrow} = [<->, line width=1.5pt, draw=gray, text=gray]

\usepackage{pifont}

\usepackage{subcaption}

\usepgfplotslibrary{groupplots,fillbetween}
\usepackage{pgfplotstable}

\usepackage{ifthen}
\usepackage{forloop}
\usepackage{listings}
\usepackage[lined,boxed,commentsnumbered,linesnumbered]{algorithm2e}

\definecolor{codegreen}{rgb}{0,0.6,0}
\definecolor{codepurple}{rgb}{0.58,0,0.82}
\definecolor{backcolour}{rgb}{0.95,0.95,0.92}
\lstdefinestyle{buzz}{
    backgroundcolor=\color{black!5},   
    commentstyle=\color{codegreen},
    keywordstyle=\color{blue},
    numberstyle=\tiny\color{black!30},
    stringstyle=\color{codepurple},
    basicstyle=\footnotesize\ttfamily,
    breakatwhitespace=false,         
    breaklines=true,                 
    captionpos=b,                    
    keepspaces=true,                 
    numbers=left,                    
    numbersep=5pt,                  
    showspaces=false,                
    showstringspaces=false,
    showtabs=false,                  
    tabsize=2,
}
\newcommand{\revision}[1]{{\color{black} #1}}
\newcommand{\secondrevision}[1]{{\color{black} #1}}

\SetAlCapSkip{0.5em}

\title{\LARGE \bf
On Your Own: Pro-level Autonomous Drone Racing\\ in Uninstrumented Arenas
}

\author{
Michael Bosello$^\dag$, Flavio Pinzarrone$^\dag$, Sara Kiade$^\dag$, Davide Aguiari$^\dag$, Yvo Keuter$^\dag$, Aaesha AlShehhi$^\dag$,\\
Gyordan Caminati$^\dag$, Kei Long Wong$^{\dag \P \S}$, Ka Seng Chou$^{\dag \P \S}$, Junaid Halepota$^\dag$, Fares Alneyadi$^\dag$,\\
Jacopo Panerati, and Giovanni Pau$^{\dag \S}$
\thanks{ 
$^\dag$Autonomous Robotics Research Center, Technology Innovation Institute, Abu Dhabi, UAE.
        E-mails:
        {\tt\footnotesize \{firstname.lastname\}@tii.ae}%
}
\thanks{ 
$^\P$Macao Polytechnic University, Macao SAR, China.
}
\thanks{ 
$^\S$University of Bologna, Bologna, Italy.
}
}

\begin{document}
\maketitle
\thispagestyle{empty}
\pagestyle{empty}

\begin{abstract}
Drone technology is proliferating in many industries, including agriculture, logistics, defense, infrastructure, and environmental monitoring. 
Vision-based autonomy is one of its key enablers, particularly for real-world applications.
This is essential for operating in novel, unstructured environments where traditional navigation methods may be unavailable.
Autonomous drone racing has become the \emph{de facto} benchmark for such systems.
State-of-the-art research has shown that autonomous systems can surpass human-level performance in racing arenas. 
However, the direct applicability to commercial and field operations is still limited, as current systems are often trained and evaluated in highly controlled environments.
In our contribution, the system's capabilities are analyzed within a controlled environment---where external tracking is available for ground-truth comparison---but also demonstrated in a challenging, uninstrumented environment---where ground-truth measurements were never available.
We show that our approach can match the performance of professional human pilots in both scenarios.
We also publicly release the data from flights carried out by a world-class human pilot: \href{github.com/tii-racing/drone-racing-dataset}{github.com/tii-racing/drone-racing-dataset}.
\\
\\
Video: \href{youtube.com/watch?v=SNw-zXgv_vA}{youtube.com/watch?v=SNw-zXgv\_vA}
\end{abstract}

\section{Introduction}
\label{sec:introduction}

Drone technology can now be found in a wide range of industries, such as agriculture, the logistics and delivery sectors, defense, infrastructure inspection, and environmental monitoring. 
For all these applications, drones can improve efficiency and operational safety.

The development of vision-based state estimation and control, in particular, is at the core of the deployment of autonomous drone systems in real-world applications.
Vision-based autonomy is essential for operating in novel, unstructured, and uninstrumented environments, where traditional navigation methods may be unreliable or unavailable~\cite{hanover2023autonomous}.

\begin{figure}
    \includegraphics{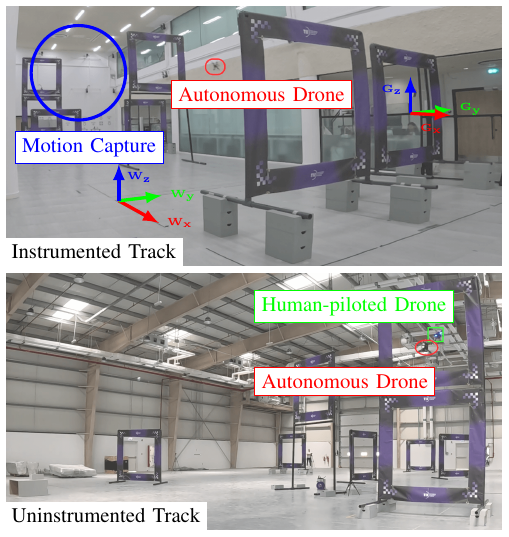}
     \caption{
    The two arenas used during the experiments, showing the autonomous drone (red) performing a lap in the instrumented one (top), and a head-to-head versus a human pilot (green) in the uninstrumented track (bottom). }
    \label{fig:frontpage}
\end{figure}

The last two decades of rapid progress in AI have seen machine learning systems challenge, and even defeat, humans in several tasks (from image recognition to the game of Go, to self-driving).
Since the 2019 AlphaPilot AI Drone Innovation Challenge~\cite{WinnerAlphapilot, scaramuzza-auro}, drone racing has emerged as the \emph{de facto} benchmark for evaluating vision-based aerial autonomy.
In 2023, state-of-the-art research has demonstrated that autonomous racing drones can surpass human-level performance in controlled racing arenas~\cite{scaramuzza-nature}.
However, the applicability of fast, vision-based autonomous flight to practical commercial operations is still limited, as current systems are typically trained and evaluated in highly controlled (often externally instrumented) environments.

The main contributions of this work are as follows. 
\begin{itemize}
    \item Achieving pro-level autonomous drone racing in both controlled environments---where external tracking is available---as well as in equally challenging, uninstrumented environments---where ground-truth measurements were never available (see Fig. \ref{fig:frontpage}).
    \item Presenting a perception and control stack for autonomous drone racing that does not require fine-tuning with ground truth for residual estimation (and has proven to be resilient to multiple lighting conditions).
    \item Releasing pro-level piloting data from the instrumented track using the same format as in~\cite{bosello2024}, adding six new flights by a world champion pilot (Sec. \ref{sec:dataset}).
\end{itemize}
\begin{table*}[h!]
    \centering
    \caption{
    Comparison of ``human v. robot'' drone racing literature
    }
\newcommand{\cmark}{\ding{51}}%
\newcommand{\xmark}{\ding{55}}%

\centering
\begin{tabular}{
>{\centering}p{.79cm}
>{\centering}p{.79cm}
>{\centering}p{1.99cm}
>{\centering}p{1.99cm}
>{\centering}p{1.19cm}
>{\centering}p{1.19cm}
>{\centering}p{1.49cm}
>{\centering}p{1.49cm}
>{\centering}p{1.99cm}
}
    \toprule
    \multirow{2}{*}{Ref.}
    & \multirow{2}{*}{Year}
    & Vehicle
    & Competition /
    & \multicolumn{2}{c}{External Sensing}
    & Human Best
    & Robot Best
    & Head-to-head
    \tabularnewline

    & %
    & Type
    & Contribution
    & On Track
    & In Race
    & Speed / Lap
    & Speed / Lap
    & Winner
    \tabularnewline
    \cmidrule(lr){1-9}

    \multirow{2}{*}{\cite{WinnerAlphapilot}} %
    & \multirow{2}{*}{2019}
    & \multirow{2}{*}{Quadcopter}
    & \multirow{2}{*}{AlphaPilot}
    & \multirow{2}{*}{\xmark}
    & \multirow{2}{*}{\xmark}
    & n/a \\ n/a
    & 9.19m/s \\ 12.00s
    & \multirow{2}{*}{n/a}
    \tabularnewline
    \cmidrule(lr){1-2} \cmidrule(lr){3-4} \cmidrule(lr){5-9}
    \multirow{2}{*}{\cite{scaramuzza-nature}} %
    & \multirow{2}{*}{2023}
    & \multirow{2}{*}{Quadcopter}
    & \multirow{2}{*}{Research}
    & \multirow{2}{*}{\cmark}
    & \multirow{2}{*}{\xmark}
    & 21.54m/s \\ 5.19s\hyperlink{hyref:01}{$^\dag$}
    & 19.44m/s \\ 5.11s\hyperlink{hyref:01}{$^\dag$}
    & \multirow{2}{*}{Robot}
    \tabularnewline
    \cmidrule(lr){1-2} \cmidrule(lr){3-4} \cmidrule(lr){5-9}
    \multirow{2}{*}{\cite{bosello2024}} %
    & \multirow{2}{*}{2024}
    & \multirow{2}{*}{Quadcopter}
    & \multirow{2}{*}{Dataset}
    & \multirow{2}{*}{\cmark}
    & \multirow{2}{*}{\cmark}
    & 9.58m/s \\ 6.04s
    & 21.83m/s \\ 3.20s
    & \multirow{2}{*}{Robot}
    \tabularnewline
    \cmidrule(lr){1-2} \cmidrule(lr){3-4} \cmidrule(lr){5-9}
    \multirow{2}{*}{\emph{This}}
    & \multirow{2}{*}{2025}
    & \multirow{2}{*}{Quadcopter}
    & \multirow{2}{*}{Research}
    & \multirow{2}{*}{\xmark}
    & \multirow{2}{*}{\xmark}
    & 25.63m/s \\ 5.04s
    & 21.15m/s \\ 5.60s
    & \multirow{2}{*}{Human}
    \tabularnewline

    \bottomrule
\end{tabular}
 \newline \hfill \newline
 {\footnotesize
 \hypertarget{hyref:01}{$^\dag$}Computed from the MoCap recordings released in the extended data of~\cite{scaramuzza-nature}, filtered by frequency to account for missing MoCap frames. Speed is computed as the derivative of position and smoothed using an exponentially weighted mean with $\alpha$ = 0.3.
 }
     \label{tab:related}
\end{table*}

\section{Related Work}
\label{sec:related}

Autonomous drone racing~\cite{hanover2023autonomous} (and the related discipline of autonomous, or self-driving, car racing) is an emerging racing sport characterized by fundamental research and technological challenges, in particular the need for fast and robust perception, planning, and control.

In car racing, since the 2004/2005 DARPA Grand Challenges, various leagues and research challenges have been established, including the Indy Autonomous Challenge (IAC) and the Abu Dhabi Autonomous Racing League (A2RL)---the latter organizing ``Man vs Machine'' events for both drone and car racing~\cite{a2rl-dr}---aimed at accelerating progress in performance, research, and entertainment.

In drone racing, competitions like the AlphaPilot AI Drone Innovation Challenge~\cite{scaramuzza-auro} marked significant milestones, with the best teams~\cite{WinnerAlphapilot} showing important advances in the use of deep learning for gate detection, Visual-Inertial Odometry (VIO), and dynamics modeling.

At the core of the control loop are often model predictive techniques (like Model Predictive Contouring Control (MPCC)~\cite{romero2022mpc} or Perception-Aware Model Predictive Control (PAMPC)~\cite{falanga2018pampc}) that explicitly consider dynamics and sensor limitations, enabling high-speed, agile flight.

Nonetheless, the latest research developments showed that the use of deep reinforcement learning for control could also make autonomous racing drones capable of outperforming human champion pilots~\cite{scaramuzza-nature, scaramuzza-scirob, ferede2025}.
Table~\ref{tab:related} summarizes the human-piloted and autonomous performance recorded in this work alongside those published in similar recent work.

Beyond competitions, research progress is also facilitated by benchmarks and datasets~\cite{bosello2024}, which typically include visual, inertial, and motion capture data from aggressive flights, crucial for developing novel methods and enabling quantitative comparisons. 

Still open research questions include: safely and effectively managing mixed human-robot multi-vehicle racing, reliable state estimation at extreme speeds, coordinating multiple autonomous racers, ensuring safety, and improving sim-to-real algorithm transfer. 

Insights from autonomous high-speed flight demonstrations, such as this one, and the analysis of pro-level human pilot data, like those released alongside this article, can offer valuable cues for future research.

\section{Drone Platform}
\label{sec:platform}

\begin{figure}
    \centering
    \includegraphics{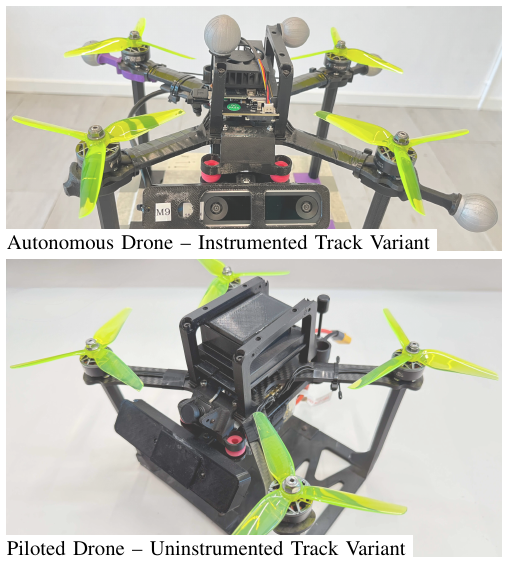}
     \caption{
    Quadrotors used in the experiments: the autonomous drone (including markers for the instrumented track, top); and the piloted drone (with replicas of the autonomy components, without markers for the uninstrumented track, bottom).
    }
    \label{fig:drone}
\end{figure}

Racing drones require a combination of agility, durability, and power. In addition to that (and unlike human-piloted drones), autonomy also demands careful integration of on-board compute with high-performance sensors. 

The components listed in this section result in an autonomous drone capable of speeds over 25m/s along an agile indoor trajectory and weighing just 665.5g (without battery).

\subsection{Hardware}
\label{sec:hardware}

The hardware components are partially based on the open-design from~\cite{bosello2024}, these include the Flight Controller (FC), motors, electronic
speed controller, and battery eliminator circuit. However, the frame and 3D-printed parts have been modified to accommodate the stereo-camera, adopting the same frame (Armattan Chameleon Ti 6'') as in~\cite{foehnscirob}.

\subsubsection{Motors and Propellers}

The quadrotor features the T-motor F60 PRO V 2020KV motors, as in~\cite{bosello2024}, paired with HQProp HeadsUp R38 propellers to accommodate the professional pilots' unanimous preference for a lower blade pitch. This setup results in a Thrust-to-Weight Ratio (TWR) of~$\sim$7 at full battery capacity, measured using a thrust bench.

\subsubsection{Camera and Sensors}

The platform is equipped with an Intel RealSense T265 stereo-camera, which captures fisheye grayscale images at 30Hz per lens and offers VIO at 200Hz. A custom damping mechanism (Fig.~\ref{fig:drone}) minimizes image blur, improving VIO accuracy during aggressive flight. Additionally, an InvenSense MPU6000 IMU embedded in the FC provides high-frequency acceleration and gyroscopic~data.

\subsubsection{Compute}

As a companion board, we use the NVIDIA Orin NX module, installed on the A603 carrier board by Seeedstudio. The Orin NX module comes with JetPack 5.1.2 (i.e., Linux Kernel 5.10, Ubuntu 20.04-based root file system, and CUDA 11.4 support). \revision{We select the \textit{MAXN} power mode to maximize core usage and clock frequencies.}

\subsubsection{FPV Replica}
Drone replicas of the autonomous model were prepared for the pilots. These replicas were outfitted with an HDZero FPV system (camera, transmitter, and antenna). The autonomy components (i.e., NVIDIA computer and Intel camera) were replaced by 3D-printed replicas filled with lead. This modification ensured that the piloted drones matched the weight and weight distribution of the autonomous one while minimizing the risk of damaging sensitive components. The autonomous and piloted drones are shown in Fig.~\ref{fig:drone}.

\subsection{Software}
\label{sec:software}

\subsubsection{Autopilot}
\label{sec:software:autopilot}
Our drone requires low-latency, real-time reactive control. This is managed by the FC, running Betaflight (BF) 4.3.2 firmware tuned for the platform~\cite{betaflight}.
The FC is interfaced using the MultiWii Serial Protocol (MSP), enabling efficient bidirectional data exchange between the FC and the onboard computer.
This allows for IMU readings at a rate up to 500Hz over a 1MBaud serial connection (much more than the 10Hz obtainable with the SBUS protocol, as reported in~\cite{foehnscirob}).
Additionally, the MSP Override feature provides the onboard computer with complete control over the flight channels.
A real-time Linux kernel setup---with priority granted to the MSP process---was crucial to achieve the performance reported in this letter.

\subsubsection{Vison and Autonomy}
Our platform uses a hybrid autonomy strategy, integrating \emph{(i)} \emph{ad hoc} perception and control modules (see Sec.~\ref{sec:autonomy}), implemented in the Robot Operating System 2 (ROS2), with \emph{(ii)} Intel's proprietary RealSense SDK (\texttt{\small librealsense2}), which provides an estimate of the camera position during the flight, albeit prone to significant drift even over very short time.

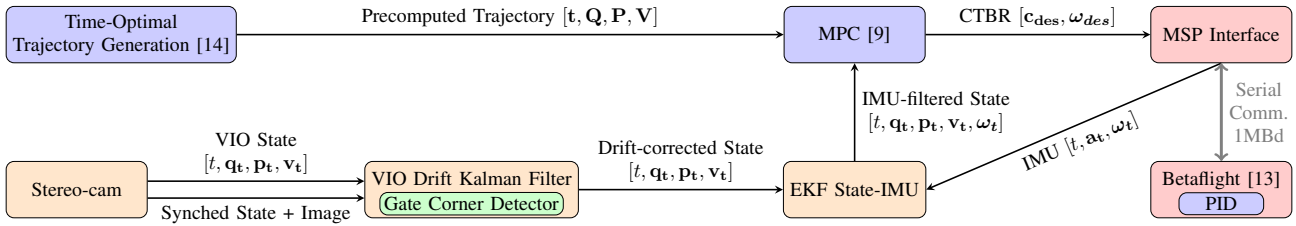
\begin{figure*}
    \centering
    \scalebox{0.75}{
\tikzexternaldisable
\begin{tikzpicture}[node distance=2.5cm]

\node (traj) [block, align=center] {Time-Optimal \\ Trajectory Generation~\cite{foehn2021CPC}};
\node (controller) [block, right of=traj, xshift=10.5cm] {MPC~\cite{falanga2018pampc}};
\node (msp) [red_block, right of=controller, xshift=4cm] {MSP Interface};
\node (stereocam) [processor, below=of traj.west, anchor=west, yshift=-0.25cm] {Stereo-cam};
\node (vio) [processor, right of=stereocam, xshift=4.5cm, align=center] {VIO Drift Kalman Filter\\ };
\node (inner) [vision, inner sep=2.2pt, yshift=-0.25cm] at (vio.center) {Gate Corner Detector};
\node (ekf) [processor, below of=controller, yshift=-0.25cm] {EKF State-IMU};
\node (bf) [red_block, below of=msp, yshift=-0.25cm, align=center] {Betaflight \cite{betaflight} \\ }; 
\node (bf_inner) [inner_block, inner sep=2.2pt, yshift=-0.25cm] at (bf.center) { PID };

\draw [arrow] (traj.east) -- node[anchor=south] {Precomputed Trajectory \([\mathbf{t}, \mathbf{Q}, \mathbf{P}, \mathbf{V}]\)} (controller.west);
\draw [arrow] (controller.east) -- node[anchor=south] {CTBR \([\mathbf{c_{des}}, \boldsymbol{\omega_{des}}]\)} (msp.west);
\draw [arrow, align=center] ([yshift=0.15cm]stereocam.east) -- node[anchor=south] {VIO State \\ \([t, \mathbf{q_t}, \mathbf{p_t}, \mathbf{v_t}]\)} ([yshift=0.15cm]vio.west);
\draw [arrow] ([yshift=-0.15cm]stereocam.east) -- node[anchor=north] {Synched State + Image} ([yshift=-0.15cm]vio.west);
\draw [arrow, align=center] (vio.east) -- node[anchor=south] {Drift-corrected State \\ \([t, \mathbf{q_t}, \mathbf{p_t}, \mathbf{v_t}]\)} (ekf.west);
\draw [arrow, align=center] (ekf.north) -- node[anchor=west] {IMU-filtered State \\ \([t, \mathbf{q_t}, \mathbf{p_t}, \mathbf{v_t}, \boldsymbol{\omega_t}]\)} (controller.south);
\draw [arrow] (msp.south) -- node[midway,below,sloped] {IMU \([t, \mathbf{a_t}, \boldsymbol{\omega_t}]\)} (ekf.east);
\draw [doublearrow, align=center] (msp.south) -- node[anchor=west] {Serial \\ Comm. \\1MBd} (bf.north);

\end{tikzpicture}
\tikzexternalenable     }
    \caption{
        Architecture diagram illustrating the data flow between the main hardware and software components,  
        \revision{including: \emph{vision} (green, Sec.~\ref{sec:vision-stack}); \emph{state estimation} (orange, Sec.~\ref{sec:state-estimation-stack}); \emph{control} (blue, Sec.~\ref{sec:control-stack}); and \emph{autopilot} (red, Sec.~\ref{sec:software:autopilot}). 
        Unlike~\cite{scaramuzza-nature}, we introduce a second-stage EKF leveraging the high-frequency FC IMU, which produces a sufficiently smooth state estimate to enable the use of classical control and MPC. 
        High-frequency readings (500\,Hz) from Betaflight are supported by our optimized MSP setup (Sec.~\ref{sec:software}), offering higher performance than the SBUS protocol adopted in~\cite{foehnscirob}.
        }
    }
    \label{fig:schematics}
\end{figure*}

\begin{figure*}
    \centering
    \includegraphics[trim={0cm 0cm 0cm 0cm},clip,width=\columnwidth*2]{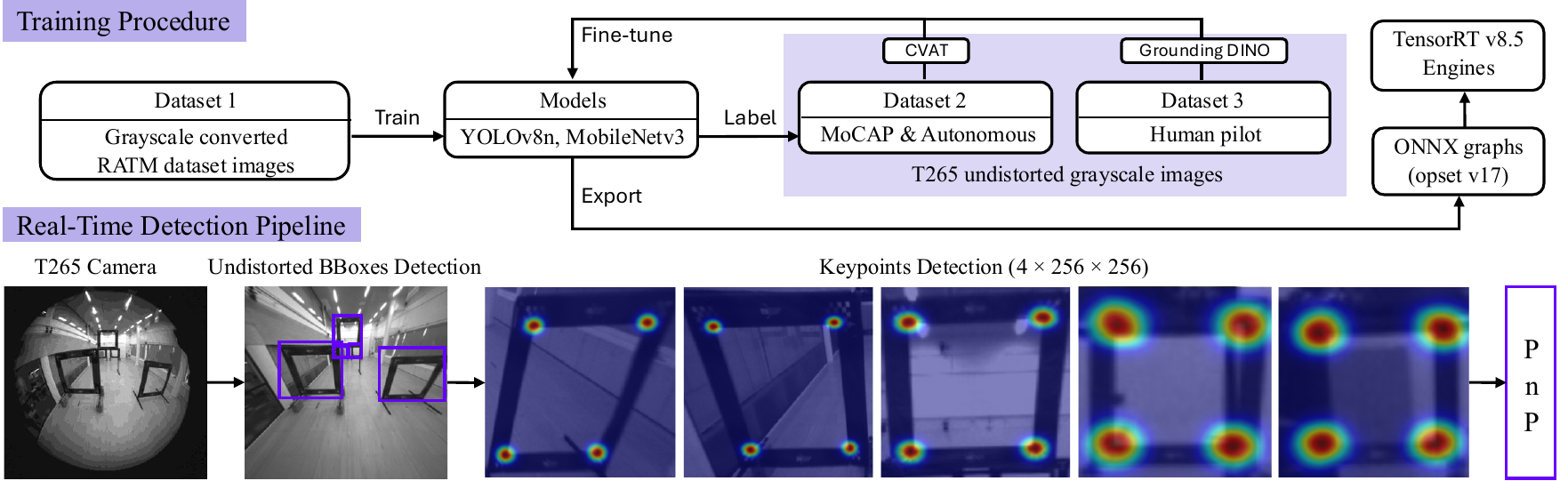}
\caption{
Vision stack schematics illustrating the training procedure (top) and the detection pipeline (bottom). 
\revision{The training procedure comprises three steps (designed to minimize human supervision while preserving accuracy): \emph{(i)} training on the RATM dataset~\cite{bosello2024}; \emph{(ii)} auto-labeling, followed by human refinement in CVAT; and \emph{(iii)} labeling track images with Grounding DINO from human examples. By the final stage, fine-tuning for peak performance in new environments requires as few as 80 human-corrected frames. 
In the detection pipeline, YOLOv8n detects gate bounding boxes, while MobileNetV3-Small~\cite{mobilenetv3} estimates their keypoints. 
Leveraging GPU acceleration (see Sec.~\ref{sec:vision-stack}), the stack achieves per-frame latency under 30\,ms on $640\times640$ images. This yields lower delay and larger input resolution compared to~\cite{scaramuzza-nature}, resulting in higher pixel-level precision and potentially improved PnP accuracy.}}
    \label{fig:vision_stack}
\end{figure*}

\section{Autonomous Drone Racing}
\label{sec:autonomy}

The autonomy stack has three main components: vision, state estimation, and control. The vision module provides gate corner detections. The state estimation uses those detections to correct the drift of the Intel T265 VIO, and the FC IMU readings for refining the state. Given the current state estimate and a time-optimal reference trajectory, the control block computes the command to be executed. A schematic of the stack is presented in Fig.~\ref{fig:schematics}.

\subsection{Vision Stack}
\label{sec:vision-stack}

The vision stack detects gates using grayscale images captured from the stereo-camera.
For this, only the image frame from the right camera is used, as its position is closer to the drone's center. 
Every camera is calibrated with the fisheye model to obtain its intrinsics. 

A gate is represented by its four inner corners. 
The pipeline includes two steps---gate detection and corner detection---each using a separate Convolutional Neural Network (CNN)-based model.
\revision{All models are converted into ONNX graphs with opset v17, and then into TensorRT (v8.5) engines, with half-precision floating-point (FP16), to exploit the Orin NX's Ampere GPU.
The measured per-frame latency of the detection stack is 24--30ms, varying with the number of detected gates and the instantaneous CPU load.}
Fig.~\ref{fig:vision_stack} illustrates the training procedure and detection flow of the vision stack.

For gate detection, we use You Only Look Once (YOLO), specifically, the YOLOv8n model with $3.2$ million parameters, one class (gates), and input size (640$\times$640). %

For corner detection, the pixels corresponding to each detected gate region are cropped and forwarded to a custom key-points detection model, MobileNetV3-Small~\cite{mobilenetv3} with $1.1$ million parameters.
The input size of the model is (256$\times$256), and its weights are initially pretrained for a classification task using the ImageNet-1K dataset. %

Following the pretraining phase, only the feature extraction layers are kept, while the classification head is replaced with a key-point head composed of five CNN layers.
The first three layers employ scaling factors of $[4, 4, 2]$, with convolutional filters of $[256, 128, 64, 32, 4]$.

The model produces heatmaps representing the confidence of four key-point coordinates, yielding an output shape of $(N, 4, 256, 256)$, where $N$ denotes the batch size (or the number of detectable gates at inference time for parallelization).
The four channels represent the four key-points,
 respectively.

\subsubsection{Data Collection and Training Procedure}
The training process consists of fine-tuning pretrained models for both tracks.
Gate and corner detection models were pretrained on the grayscale version of the dataset in~\cite{bosello2024} and fine-tuned on undistorted images using the calibrated intrinsics.

The instrumented track fine-tuning dataset comprises 3,412 frames from autonomous flights.
These are initially labeled by the aforementioned pretrained models, then manually corrected using the annotation software CVAT\footnote{\href{https://www.cvat.ai/}{https://www.cvat.ai/}}.

For the uninstrumented track, we use model distillation to make up for the lack of existing labeled data.
A foundational model, Grounding DINO~\cite{mmdetection}, is fine-tuned using 80 manually labeled frames from human-piloted flights, then used to generate gate detection labels for the entire flights, which are used to fine-tune YOLO.
For corner detection, the instrumented track model is applied for auto-labeling, 
with frames exhibiting detection errors manually corrected for fine-tuning.

\subsection{State Estimation Stack}
\label{sec:state-estimation-stack}
Our state estimation stack leverages VIO from the Intel T265.
Although VIO provides highly accurate short-term estimates, its performance degrades over time, leading to significant drift, especially during fast maneuvers that introduce motion blur. 
To mitigate this drift, we incorporate detections of the only predefined landmarks in drone racing: the gates. By identifying the corners of the gates in the captured images, the drone's pose relative to the gate is computed by solving the Perspective-n-Point (PnP) problem, \secondrevision{of which only the positional component is used.
} The global \secondrevision{position} of the drone is then refined using the gate's known location in the track layout, \secondrevision{whereas the attitude is left uncorrected, as it exhibits a substantially lower drift over time.}

Using fixed exposure and carefully tuning the exposure time and gain, we obtain good image brightness with minimal motion blur, ensuring reliable VIO. 
\revision{We enable ``mapping'', allowing the device to build and update an internal map for loop closure and small drift correction. In contrast, we disable ``relocalization'', which attempts to re-align the estimate after large drifts, and ``pose jumping'', which can cause discontinuous estimates. This configuration provided the best balance of VIO accuracy and reliability, while still allowing us to correct larger drifts using gate measurements.} The right camera operates at 30Hz, and its frames are synchronized with the corresponding VIO estimates.

\revision{
Our drift correction strategy starts with the detection of all visible gates in the undistorted image (see Sec.~\ref{sec:vision-stack} and Fig.~\ref{fig:vision_stack}). Then, we solve a PnP problem for each detected gate with four visible corners using OpenCV's \texttt{\small SOLVEPNP\_ITERATIVE}, which initializes the solution using homography decomposition, as the gates are planar, and refines it with a nonlinear Levenberg-Marquardt minimization scheme. Using these gate measurements and the known track map, we set up a Kalman Filter similar to the one presented in~\cite{scaramuzza-nature} to estimate the translational drift of the VIO. In our case, the state vector is \(\mathbf{x} = \mathbf{p}_d^T \in \mathbb{R}^3\), representing the positional drift vector. Unlike~\cite{scaramuzza-nature}, we do not estimate the drift velocity as it has a less stable behaviour when loop closure is performed at the camera firmware level. Therefore, the state \(\mathbf{x}\) and covariance \(\mathbf{P}\) are propagated according to: 
\begin{equation}
    \mathbf{x}_{k+1} = F\mathbf{x}_k\quad\quad\quad
    P_{k+1} = FP_kF^T + Q
    \label{eq:KF_state_cov}
\end{equation}
\begin{equation}
    F = \mathbb{I}^{3\times3}\quad\quad\quad
    Q = \mathbb{I}^{3\times3}\frac{1}{4}\mathbf{d}\mathbf{\textit{t}}^4\sigma_a^2
    \label{eq:KF_trans_proc_noise}
\end{equation}
The filter state and covariance are initialized to zero, while the process noise is set to \(\sigma_a^2 = 8\). The filter is then updated whenever a new measurement \(\mathbf{z}_k\) (a \secondrevision{position} estimate from a gate detection) is available, using the Kalman filter equations:
\begin{equation}
    K_k = P_k^-H_k^T(H_kP_k^-H_k^T+R)^{-1}
    \label{eq:K_gain}
\end{equation}
\begin{equation}
    \mathbf{x}_k^+ = \mathbf{x}_k^- + K_k(\mathbf{z}_k - H(\mathbf{x}_k^-))
    \label{eq:state_update}
\end{equation}
\begin{equation}
    P_k^+ = (I - K_kH_k)P_k^-
    \label{eq:state_cov_update}
\end{equation}
in which \(K_k\) is the Kalman gain, \(R\) is the measurement covariance matrix, estimated via Monte-Carlo sampling as described in~\cite{scaramuzza-nature}, and \(H_k\) is the measurement matrix.
When several gates are detected in a single camera frame, all relative pose estimates are stacked and processed in the same Kalman filter update step \secondrevision{as multiple simultaneous measurements}. The resulting drift estimate is constantly used to correct the VIO position estimate. As a final step, the state estimate is refined by fusing it with IMU data from the FC, which operates at 500Hz. This fusion is performed using an Extended Kalman Filter (EKF) with state vector  \(\mathbf{x} = [\mathbf{q}^T, \mathbf{p}^T, \mathbf{v}^T, \mathbf{b}_{\omega}^T, \mathbf{b}_a^T]^T \in \mathbb{R}^{16}\) and error state vector \(\mathbf{e} = [\mathbf{\psi}, \mathbf{\theta}, \mathbf{\phi}, \mathbf{p}^T]^T \in \mathbb{R}^{6}\), in which \(\mathbf{q}\) represents the drone's orientation quaternion and \(\mathbf{\psi}, \mathbf{\theta}, \mathbf{\phi}\) its Euler angles representation, \(\mathbf{p}\) is its position, \(\mathbf{v}\) is its linear velocity, \(\mathbf{b}_{\omega}\) is the bias of the gyroscope and \(\mathbf{b}_a\) is the bias of the accelerometer. The propagation step follows the equations:
\begin{equation}
    \mathbf{x}_{k+1} = F\mathbf{x}_k\quad\quad\quad
    P_{k+1} = FP_kF^T + WQW^T
    \label{eq:KF_state_p}
\end{equation}
Where \(\mathbf{P}\) is the state covariance matrix, \(\mathbf{F}\) is the state transition matrix, and \(\mathbf{W}\) is the Jacobian of the error-state transition model with respect to process noise. The update step is then defined as in \ref{eq:K_gain}, \ref{eq:state_update}, \ref{eq:state_cov_update}, but in this case  \secondrevision{the measurement is \(\mathbf{z}_k = [\mathbf{\psi}, \mathbf{\theta}, \mathbf{\phi}, \mathbf{p}^T]^T \in \mathbb{R}^{6}\), representing attitude and gate-corrected position estimates from VIO, and} \(\mathbf{R}\) is a static diagonal covariance matrix.
}

\subsection{Control Stack}
\label{sec:control-stack}

Our control stack couples Model Predictive Control (MPC) with a time-optimal trajectory generator.

\subsubsection{Time Optimal Trajectory Generation}
\label{sec:trajgen}
An open-source time-optimal trajectory generator~\cite{foehn2021CPC} was utilized to create reference trajectories for both tracks. The generator calculates trajectories minimizing the time required to reach specified waypoints, taking into account \revision{the full rigid-body} drone dynamics and actuator constraints. \revision{The linear aerodynamic drag coefficients were not included.}

Each gate was assigned two waypoints, positioned at the center of the gate along the y- and z-axes. Along the x-axis, the waypoints were placed at -0.4m and +0.4m (-0.4m and +1.25m for the Split-S). \revision{This displacement is defined in gate-frame, and it is transformed in world-frame using the known gate yaw angle (in radians) specified in Fig.~\ref{fig:tracks}. The world- and gate-reference frame are shown in Fig.~\ref{fig:frontpage}.
\begin{equation}
\begin{cases}
x_{wp} = x_{gate} \;\pm\; 0.4 \times \cos(\theta_{gate}) \\
y_{wp} = y_{gate} \;\pm\; 0.4 \times \sin(\theta_{gate})
\end{cases}
\end{equation}

} We center the waypoints in the y- and z-axes for caution and use double waypoints along the x-axis to ensure that the optimized trajectory does not cut through the gate banners.
To ensure the feasibility of the trajectories, we set a (conservative) TWR ratio of 3.8 in the generation parameters to mitigate potential modeling errors.

\subsubsection{MPC Problem}
\label{sec:mpc}
The open-source MPC framework from~\cite{falanga2018pampc} served as the foundation of our implementation.
However, the perception-aware objectives were disabled, and the remaining cost weights were carefully tuned, prioritizing precise trajectory tracking and robustness against noisy state estimation.
To compensate for the command delay\revision{---including MPC computation, FC communication, and motor actuation delay---}we incorporated a state predictor. This predictor estimates the current state based on \revision{point-mass model} dynamics and previous commands, compensating for delays before passing the information to the MPC controller, which then computes the optimal control inputs. This modification was crucial to ensure proper trajectory execution.

\subsubsection{Betaflight PID}

The MPC setpoints are defined as Collective Thrust and Body Rates (CTBR), which must be converted into PWM rotor signals to control the drone. This conversion is handled by the internal PID controllers of BF, which utilize gyro feedback to achieve the desired setpoints. The CTBR commands can be directly mapped to BF channels and transmitted to the FC. The PID gains in BF were tuned to optimize the drone’s autonomous performance.

\begin{figure}
    \centering
    \includegraphics{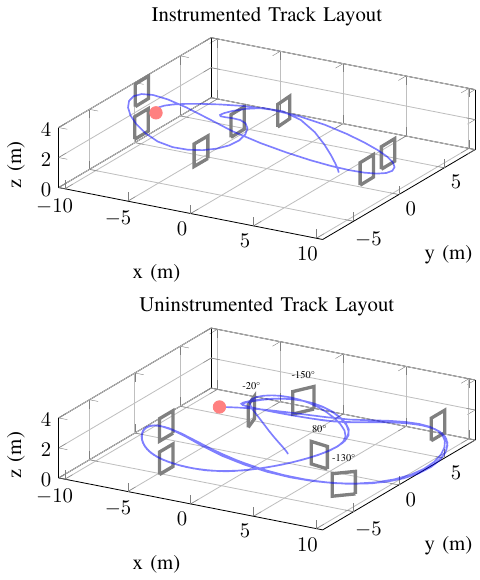}
     \caption{
    Gate setup for the instrumented (Sec.~\ref{sec:instrumented-track}, top) and uninstrumented (Sec.~\ref{sec:uninstrumented-track}, bottom) tracks, including starting point (in red), and the time-optimal reference trajectories (in blue). \revision{Yaw is reported for gates with non-zero yaw.}
    }
    \label{fig:tracks}
\end{figure}

\section{Racing Tracks}
\label{sec:tracks}

For our experiments, we used two different tracks (Fig. \ref{fig:tracks}). The first track was set up in a flight arena equipped with a motion capture (MoCap) system, providing ground-truth data to analyze speed, gate-to-gate times, traveled distance, state estimation, and trajectory tracking. The second track was prepared in a larger, uninstrumented hangar. In this case, quantitative evaluation was limited to lap \revision{and sector} times.

The tracks were created using gates made of PVC pipes and covered with fabric banners featuring printed designs. Each gate measures 7-by-7ft (213.36cm) with an inner opening of 5-by-5ft (152.4cm), matching closely the standards set by prominent drone racing leagues\footnote{\href{https://www.multigp.com/product/drone-racing-gate-bundle}{https://www.multigp.com/product/drone-racing-gate-bundle}}.

\subsection{Instrumented Track}
\label{sec:instrumented-track}

\subsubsection{Technical Specifications}
The \emph{Track RATM} from~\cite{bosello2024} was recreated in an indoor arena measuring 25 (L) by 9.7 (W) by 7 (H) meters. The MoCap system was used for the exact placement of the gates. The track features a sharp hairpin, a spiral segment, and a Split-S maneuver.
The gate poses were taken from the public 
{GitHub repository} 
accompanying~\cite{bosello2024}.

\subsubsection{Motion Capture}
The instrumented arena is equipped with a 32-camera Arqus A12 Qualisys MoCap system\footnote{\href{https://www.qualisys.com/cameras/arqus}{https://www.qualisys.com/cameras/arqus}}, capable of tracking the 6DoF poses of rigid bodies with millimeter-level precision at 275Hz. The drone presented in Sec.~\ref{sec:hardware} was outfitted with five 25mm markers, three on 3D-printed arm extensions, and two near the center of the frame. 

\subsection{Uninstrumented Track}
\label{sec:uninstrumented-track}

\subsubsection{Technical Specifications}
In the larger, uninstrumented hangar, we reproduced, as faithfully as possible, the \emph{Track Split-S} from~\cite{scaramuzza-nature}. We positioned the gates with the support of a total station, resulting in position errors in the order of a few centimeters. The gates' orientations had errors of at most ten degrees.
We should also note that the bottom gate of the Split-S in our track is 20cm lower compared to the original track due to a different gate interlocking system. 
The gates' placement is similar to that of \emph{Track RATM}.
Both tracks feature the Split-S maneuver and similar z-axis variations, although this track has softer curves and lacks the sharp hairpin and spiral segment.
The gate poses of the original track are published in the supplementary material of~\cite{scaramuzza-nature}.

\begin{table*}[t]
    \centering
    \caption{
    Summary of the $^\dag$instrumented and $^\ddag$uninstrumented flights
    }
    {
\scalebox{0.875}{
\begin{tabular}{ 
c
c
c
c
c
c
c
c
c
c
c
c
c
c
c
c
} 

\toprule

\multirow{2}{*}{Pilot} & 
\multicolumn{4}{c}{Lap Time (s)} &
\multicolumn{4}{c}{Top Speed (m/s)} &
\multicolumn{4}{c}{Path Length (m)} &
\multirow{2}{*}{Laps} &
Laps / &
\multirow{2}{*}{Crashes}
\tabularnewline 

& 
Avg. &
Std. &
Min. &
Max. &
Avg. &
Std. &
Min. &
Max. &
Avg. &
Std. &
Min. &
Max. &
&
Batteries &
 
\tabularnewline 

\cmidrule(lr){2-5}
\cmidrule(lr){6-9}
\cmidrule(lr){10-13}
\cmidrule(lr){14-14}
\cmidrule(lr){15-15}
\cmidrule(lr){16-16}

$^\dag$\texttt{\small Star23467} &
$7.71$ & 
$0.89$ & 
$6.76$ & 
$10.74$ & 
$11.90$ & 
$0.97$ & 
$9.13$ & 
$13.8$ & 
$55.68$ & 
$1.74$ & 
$51.65$ & 
$62.60$ & 
$49$ & 
$6.12$ & 
$2$ 
\tabularnewline 

$^\dag$\texttt{\small Ion FPV} &
$6.51$ & 
$0.60$ & 
$5.70$ & 
$9.04$ & 
$15.83$ & 
$1.57$ & 
$12.12$ & 
$18.51$ & 
$57.37$ & 
$1.93$ & 
$52.27$ & 
$62.85$ & 
$78$ & 
$6.5$ & 
$7$ 
\tabularnewline 

$^\dag$\texttt{\small MCK} &
$4.71$ & 
$1.25$ & 
$\mathbf{3.84}$ & 
$15.75$ & 
$20.87$ & 
$2.95$ & 
$13.16$ & 
$\mathbf{24.96}$ & 
$51.29$ & 
$5.87$ & 
$47.38$ & 
$124.68$ & 
$\mathbf{196}$ & 
$\mathbf{6.75}$ & 
$5$ 
\tabularnewline 

$^\dag$\emph{Ours} MoCap &
$\mathbf{4.44}$ & 
$\mathbf{0.11}$ & 
$4.39$ & 
$\mathbf{4.85}$ & 
$\mathbf{21.92}$ & 
$\mathbf{0.50}$ & 
$\mathbf{20.06}$ & 
$22.28$ & 
$\mathbf{47.42}$ & 
$\mathbf{0.40}$ & 
$47.15$ & 
$\mathbf{48.90}$ & 
$21$ & 
$1$ & 
$\mathbf{0}$ 
\tabularnewline 

$^\dag$\emph{Ours} VIO &
$4.65$ & 
$0.22$ & 
$4.40$ & 
$\mathbf{4.85}$ & 
$20.98$ & 
$1.29$ & 
$19.61$ & 
$22.2$ & 
$48.93$ & 
$1.11$ & 
$\mathbf{47.11}$ & 
$50.01$ & 
$6$ & 
$1$ & 
$\mathbf{0}$ 
\tabularnewline

\midrule

$^\ddag$\texttt{\small MCK} &
$\mathbf{5.80}$ & 
$0.40$ & 
$\mathbf{5.05}$ & 
$6.89$ & 
-- & 
-- & 
-- & 
-- & 
-- & 
-- & 
-- & 
-- & 
$\mathbf{63}$ & 
$\mathbf{6.89}$ & 
$\mathbf{2}$ 
\tabularnewline 

$^\ddag$\emph{Ours} ``On Its Own'' &
$6.02$ & 
$\mathbf{0.06}$ & 
$5.92$ & 
$\mathbf{6.13}$ & 
-- & 
-- & 
-- & 
-- & 
-- & 
-- & 
-- & 
-- & 
$45$ & 
$3$ & 
$4$ 
\tabularnewline 

\bottomrule

\end{tabular}
}
     }
    \label{tab:summary}
\end{table*}

\begin{figure*}
    \centering
    \includegraphics{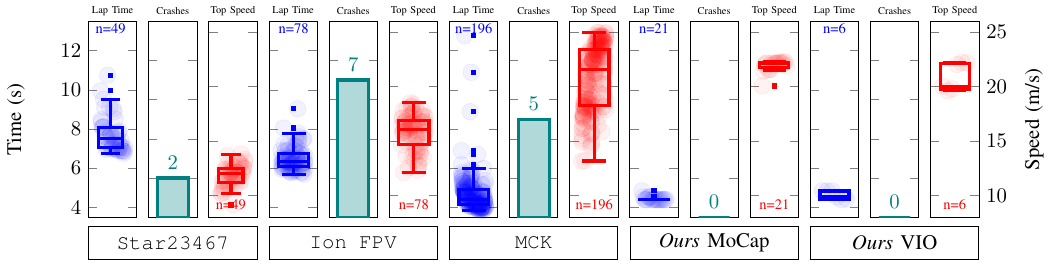}
     \caption{
    Results from the instrumented track showing the lap time distributions, the number of crashes, and the top speed distributions for the three pro-level pilots (\texttt{\small Star23467}, \texttt{\small Ion FPV}, and \texttt{\small MCK}), our autonomous system supported by external sensing (\emph{via} motion capture), and our autonomous system using only onboard sensing (VIO).
    }
    \label{fig:instrumented}
\end{figure*}

\section{Instrumented Results}
\label{sec:instrumented}

The experiments on the instrumented track (Sec.~\ref{sec:instrumented-track}), where the autonomy components were initially developed, were designed to first compare the performance of the autonomous drone with that of human pilots in an environment where more comprehensive data and statistics could be collected.
Both the autonomous and piloted drones used in these flights were equipped with MoCap markers.

\subsection{Participants}
\label{sec:participants-1}

Three professional pilots participated in the experiment, divided into two sessions. The first session included two top-ten-ranked pro-pilots from the Drone Champions League (DCL), namely Thomas Kund (\texttt{\small Star23467}) and Krutharth M. C. (\texttt{\small Ion FPV}). The second session featured a world-champion pilot, Minchan Kim (\texttt{\small MCK}), who had earned multiple titles in recent years---in MultiGP Championship, FAI World Drone Racing Championship (WDRC), and Drone Racing League (DRL)---and led his team to victory in the DCL.

\subsection{Runs}
\label{sec:runs-1}

In the first session, the two professional pilots had five days to train and record data on the instrumented track. The second session took place four months later, during which the champion pilot spent a total of five days for both (instrumented and uninstrumented) experiments, dedicating three of those days to the instrumented track.

The pilots were encouraged to train as much as needed to familiarize themselves with the platform and track, reaching their desired peak performance before starting to record timed laps for the experiment. During the training period, they also fine-tuned the Betaflight parameters to suit their preferences. While the flight hardware was fixed, the pilots had full control over the firmware configuration. They were allowed to use their preferred transmitter, cameras, personal radio controllers, and goggles to ensure optimal performance.

The pilots were also allowed to choose how many batteries or hours of practice to use each day, as well as the number of laps to complete with each battery. The autonomous drone always ran one lap per battery instead.

\subsection{Performance}
\label{sec:performance-1}

The instrumented evaluation allowed the comprehensive analysis of multiple quantitative metrics, including lap times, average speeds, and path lengths. We recorded data for three human pilots and our autonomous system, the latter using both external state estimation (MoCap) as well as fully on-board autonomy (VIO).
The results of these flights are presented in Table~\ref{tab:summary} and Fig.~\ref{fig:instrumented}.
The autonomous systems (MoCap and VIO) surpassed the humans in \emph{average} lap time and top speed, although they fell short to one out of the three pilots in terms of best individual results (achieved by \texttt{\small MCK}).
In the instrumented arena, the autonomous system showed a clear reliability advantage, completing all runs without any crash, whereas the human pilots suffered multiple incidents.

\subsection{Dataset}
\label{sec:dataset}
Six flights by the fastest pilot, \texttt{\small MCK}, totaling 240.77s of flying time and 2342.98m of traveled distance, at a top speed of 21.29m/s,
are added to our \href{https://github.com/tii-racing/drone-racing-dataset}{\texttt{\small drone-racing-dataset}}, using the format described in~\cite{bosello2024}.

\begin{figure}
    \centering
    \includegraphics{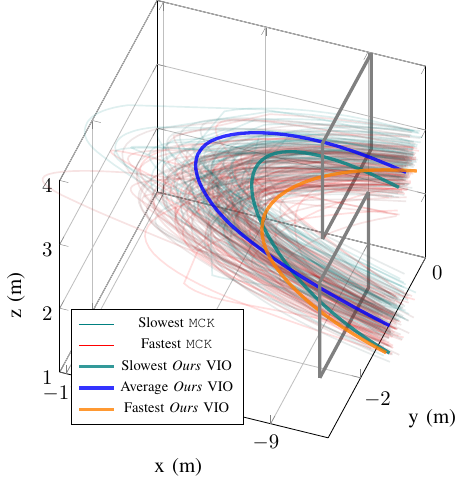}
     \caption{
    Comparison of the trajectories followed during the Split-S maneuver by the fastest pilot (\texttt{\small MCK}) and our autonomous system using only onboard sensing (VIO), sorted by lap time, on the instrumented track.
    }
    \label{fig:trajectories}
\end{figure}
\begin{figure*}
    \includegraphics{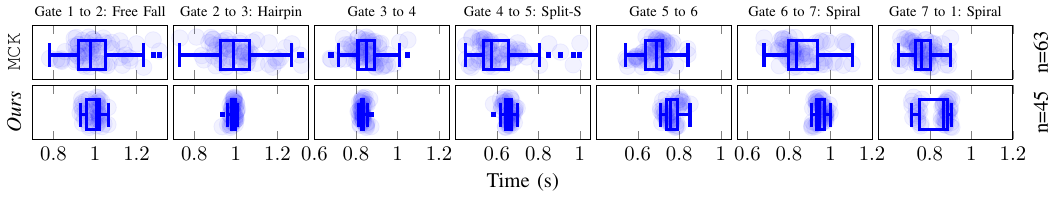}

     \caption{
    {
    Distribution of the sector times of \texttt{\small MCK} (top) and our autonomous drone (bottom) on the uninstrumented track.
    }
    }
    \label{fig:uninstrumented_sectors}
\end{figure*}

\begin{figure}
    \includegraphics{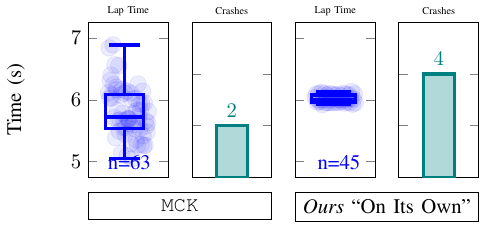}
     \caption{
    Lap time distributions and crashes for the fastest pilot
    and our autonomous system on the uninstrumented track.
    }
    \label{fig:uninstrumented}
\end{figure}

\section{Uninstrumented Results}
\label{sec:uninstrumented}

The experiments on the uninstrumented track (Sec.~\ref{sec:uninstrumented-track}) demonstrate the performance of our autonomous drone when deployed in a new environment where ground-truth data could not be recorded. The systems under scrutiny were the state estimation (Sec.~\ref{sec:state-estimation-stack}) and control (Sec.~\ref{sec:control-stack}) modules, as they are the ones that can benefit from fine-tuning using ground-truth data.

The scenario was not entirely unknown, as the basic data required for the vision stack were available.
To support the vision module, images of the new location were captured and used to fine-tune the corner detection algorithm. This adjustment allowed for accommodating the significant variations in lighting conditions caused by sunlight streaming through the hangar windows (as tests had to be carried out at all hours of the day).
The gate map, essential to support the state estimation module, was created using a total station. Although accurate, this method was considerably less precise than the map generated by the MoCap system.

\subsection{Participants}
\label{sec:participants-2}

This uninstrumented track experiment involved only the fastest world-class pilot, Minchan Kim (\texttt{\small MCK}), who competed against our autonomous drone using only on-board sensing and without ground-truth data fine-tuning (``on its own'').

\subsection{Runs}
\label{sec:runs-2}

This experiment was carried out immediately after the second session of the instrumented tests and spanned two days. It included both time trial flights and head-to-head competitions.
The time trial flights were similar to the instrumented tests, although we could only record timings. In the head-to-head flights, the autonomous and piloted drones competed simultaneously over three consecutive laps of the track. The start of these races was signaled by an audio countdown.
In both scenarios, our analysis was conducted on the basis of lap times. Moreover, the head-to-head flights provided additional insight into the endurance and resilience of the autonomous drone.
To measure lap \revision{and sector} times, we employed GoPro cameras filming at 240fps.

\subsection{Performance}
\label{sec:performance-2}

\texttt{\small MCK} achieved both a lower average and best lap time (see Table~\ref{tab:summary}, Fig.~\ref{fig:uninstrumented}).
The autonomous drone, which was 1.27\% faster than \texttt{\small MCK} on average lap time in the instrumented track, gave up a 3.65\% margin to the fastest human. Note that the second and third ranked pro-pilots were $\sim$40-60\% slower than \texttt{\small MCK} in the instrumented track.
\texttt{\small MCK} had 2 crashes, whereas the autonomous system crashed 4 times---3 of which were due to collisions with the human pilot, who was able to recover in two of these cases.
\section{Discussion}
\label{sec:discussion}

State-of-the-art research has shown that autonomous drone racing can outperform professional pilots in controlled, instrumented arenas~\cite{scaramuzza-nature}.
We set ourselves out to answer the question of whether such a level of performance could also be achieved in uninstrumented settings---as this will be crucial for the real-world deployment of these methods.

Our results were positive and showed pro-level autonomous performance against a champion pilot (\texttt{\small MCK}), over multiple laps (3), in an uninstrumented arena, with varying lighting conditions.
We should also note that the autonomous performance, even in the uninstrumented arena, albeit not always the fastest, was always remarkably more consistent than the humans (``\revision{robots [...]} are not weary'').

\secondrevision{The design choices in Sec.~\ref{sec:autonomy} and their rigorous integration were critical to achieve the demonstrated champion-level performance. Notably, the smooth state estimate produced by the dual-stage filtering (Sec.~\ref{sec:state-estimation-stack}) enabled the aggressive MPC tuning required for lap-time optimization; until the introduction of such filtering, the resulting state discontinuities had led to instability and crashes. Furthermore, the state predictor used for delay compensation (Sec.~\ref{sec:mpc}) unlocked 3D maneuvers at more sustained speed. Together, these improvements helped us cross the boundary between achieving pro-level performance and champion-level performance, corresponding to an advantage of approximately 2 seconds on the reference tracks.}

In the instrumented arena experiments (Fig.~\ref{fig:instrumented}), we observed different critical maneuvers for the human pilots and the autonomous drone. \texttt{\small Star23467} and \texttt{\small Ion FPV} reported the Split-S (Fig.~\ref{fig:trajectories}) maneuver as problematic for human pilots due to the constrained physical space, as the Split-S was placed at the end of the arena, near a wall. All pilots consistently identified the spiral section of the track as a pivotal segment for performance. For the autonomous system, the hairpin turn between gates 2 and 3 was a major limiting factor, as it became a failure point when trying to execute trajectories generated with a less conservative TWR.

In the uninstrumented arena (Fig.~\ref{fig:uninstrumented_sectors},~\ref{fig:uninstrumented}), the crashes in the head-to-head competition between \texttt{\small MCK} and the autonomous drone highlighted the autonomous system’s consistency (even to a fault) and, on the other hand, the human pilot’s greater adaptability. The autonomous system remained much more vulnerable to shared-track interactions. This type of ``human v. robot'' competition still offers many open research questions that will need to be answered to create robots that can be safely deployed alongside humans.

\revision{As a further demonstration of skill transferability, our autonomous system was also deployed in two additional uninstrumented venues---featuring novel race tracks---with only minimal tuning and limited on-site testing. Specifically, we only used 80 additional images for fine-tuning (Fig.~\ref{fig:vision_stack}), we made adjustments to the camera exposure and gain (Sec.~\ref{sec:state-estimation-stack}), and we generated new time-optimal trajectories (Sec.~\ref{sec:trajgen}). These deployments were showcased in public demonstrations at IROS~2024 and the 2024 Abu Dhabi F1 Grand Prix, where the autonomous drone outperformed the professional pilot \texttt{\small Ion FPV} in both time-trial and head-to-head races. Notably, the F1 track posed the additional challenge of outdoor 
lighting conditions. Videos of both demonstrations are included in the multimedia material.} %

\section{Conclusions}
\label{sec:conclusions}

In this letter, we presented a drone racing autonomy stack---comprising vision, state estimation, and control modules---that is capable of pro-pilot performance in uninstrumented arenas, i.e., without access to ground-truth data to fine-tune the state estimation and control modules.
In particular, we showed that our drone outperformed professional pilots in an instrumented setting and was competitive against a champion pilot in the uninstrumented one. 
In our future work, we will replace the stereo-camera with a monocular one (bringing our stack closer to its human counterpart).

\emph{Acknowledgments}---We thank the pilots, Thomas Kund (\texttt{\small Star23467}), Krutharth M. C. (\texttt{\small Ion FPV}), and Minchan Kim (\texttt{\small MCK}), for their insight and invaluable suggestions.

\bibliographystyle{./IEEEtranBST/IEEEtran}
\bibliography{./IEEEtranBST/IEEEabrv,biblio}

\end{document}